\documentclass[letterpaper]{article} % DO NOT CHANGE THIS
\usepackage[]{aaai24}  % DO NOT CHANGE THIS
\usepackage{times}  % DO NOT CHANGE THIS
\usepackage{helvet}  % DO NOT CHANGE THIS
\usepackage{courier}  % DO NOT CHANGE THIS
\usepackage[hyphens]{url}  % DO NOT CHANGE THIS
\usepackage{graphicx} % DO NOT CHANGE THIS
\usepackage{natbib}  % DO NOT CHANGE THIS AND DO NOT ADD ANY OPTIONS TO IT
\usepackage{caption} % DO NOT CHANGE THIS AND DO NOT ADD ANY OPTIONS TO IT
\usepackage{algorithm}
\usepackage{algorithmicx}
\usepackage{algpseudocode}

\usepackage{subcaption}
\usepackage{newfloat}
\usepackage{listings}
\DeclareCaptionStyle{ruled}{labelfont=normalfont,labelsep=colon,strut=off} % DO NOT CHANGE THIS
\floatstyle{ruled}
\newfloat{listing}{tb}{lst}{}
\floatname{listing}{Listing}
\newcommand{\primarykeywords}{
  (4) Theory;
  (8) Knowledge Representation/Engineering
}

\usepackage[switch, modulo]{lineno}
\usepackage{booktabs}       % professional-quality tables
\usepackage{enumerate}      % better lists
\usepackage{xltabular}
\usepackage{amsfonts}       % blackboard math symbols
\usepackage{nicefrac}       % compact symbols for 1/2, etc.
\usepackage[dvipsnames]{xcolor}  % colors
\definecolor{lightgrey}{gray}{0.8}
\usepackage{xspace}         % proper spacing after macros

\usepackage{amsmath,amsthm,amssymb}
\newtheorem{definition}{Definition}
\newtheorem{theorem}[definition]{Theorem}
\newtheorem{example}[definition]{Example}
\newcommand{\eex}{{\small$\blacksquare$}}

\usepackage{tcolorbox}
\tcbuselibrary{listings, breakable, skins}
\tcbset{colframe=black!60!white,colback=black!5!white,left=0mm,top=0.5mm,bottom=0.5mm,right=1.5mm}
\newcounter{example}
\newenvironment{boxed-example}[1][\unskip]{%
  \refstepcounter{example}
  \begin{tcolorbox}[colframe=black!60,colback=black!5!white,breakable,enhanced,title=\textbf{Example \theexample: #1}]\footnotesize\allowdisplaybreaks}{%
  \end{tcolorbox}}
\usepackage{tikz}
\usetikzlibrary{calc}
\usetikzlibrary{arrows}

\colorlet{alert}{red!80!black}

\newcommand{\Omit}[1]{}

\newcommand{\tup}[1]{\ensuremath{\langle #1 \rangle}}
\renewcommand{\O}{\mathcal{O}}

\newcommand{\iw}[1]{\ensuremath{\text{IW}(#1)}\xspace}

\newcommand{\siwR}{\ensuremath{\text{SIW}_{\text R}}\xspace}
\newcommand{\siwRx}{\ensuremath{\text{SIW}^*_{\text R}}\xspace}
\newcommand{\siwM}{\ensuremath{\text{SIW}_{\text M}}\xspace}
\newcommand{\sieve}{\textsc{Sieve}\xspace}

\newcommand{\pplus}{\hspace{-.05em}\raisebox{.15ex}{\footnotesize$\uparrow$}}
\newcommand{\mminus}{\hspace{-.05em}\raisebox{.15ex}{\footnotesize$\downarrow$}}

\newcommand{\EQ}[1]{\ensuremath{#1{\,{=}\,}0}}
\newcommand{\GT}[1]{\ensuremath{#1{\,{>}\,}0}}
\newcommand{\DEC}[1]{\ensuremath{#1\mminus}}
\newcommand{\INC}[1]{\ensuremath{#1\pplus}}
\newcommand{\UNK}[1]{\ensuremath{#1?}}

\newcommand{\prule}[2]{\ensuremath{\{ #1 \} \mapsto \{ #2 \}}}
\newcommand{\nrule}[4]{\ensuremath{#1 \,\|\, \{ #2 \} \mapsto \{ #3 \} \,\|\, #4}}
\newcommand{\xrule}[2]{\ensuremath{#1 \mapsto #2}}

\newcommand{\Obj}{\text{Obj}}

\newcommand{\M}{\ensuremath{M}\xspace}

\newcommand{\Reg}{\ensuremath{\mathfrak{R}}\xspace}
\newcommand{\preg}{\ensuremath{\mathfrak{r}}\xspace}
\newcommand{\reg}[1]{\ensuremath{\preg_{#1}}\xspace}
\renewcommand{\v}{\ensuremath{\boldsymbol{v}}\xspace}
\newcommand{\Yields}{\rightarrow^*}
\newcommand{\Reduct}[2]{\ensuremath{#1 \Yields #2}}
\newcommand{\pLoad}{\textit{Load}}
\newcommand{\Load}[2]{\ensuremath{\pLoad(#1,#2)}\xspace}

\newcommand{\module}[1]{\ensuremath{\texttt{#1}}}
\newcommand{\modname}[2][]{\ensuremath{\texttt{#2}\ifthenelse{\equal{#1}{}}{}{(\tup{#1})}}\xspace}
\newcommand{\args}{\ensuremath{\text{args}}\xspace}

\newcommand{\call}[5]{\ensuremath{(#1,#2) \mapsto (#3(#4),#5)}\xspace}

\newcommand{\dcall}[5]{\ensuremath{#1 \,\|\, \{#2\} \mapsto #3(#4) \,\|\, #5}\xspace}

\newcommand{\piclear}{\ensuremath{\pi_{\textit{clear}^*}}\xspace}
\newcommand{\pion}{\ensuremath{\pi_{\textit{on}}}\xspace}

\newcommand{\concept}[1]{\textsf{\footnotesize #1}}

\newcommand{\C}{\concept{C}\xspace}
\newcommand{\X}{\concept{X}\xspace}
\newcommand{\Y}{\concept{Y}\xspace}

\newcommand{\role}[1]{\concept{#1}}
\newcommand{\R}{\role{R}\xspace}

\newcommand{\ON}{\textit{On}\xspace}

\newcommand{\closure}[1]{\ensuremath{#1^{\circ}}\xspace}

\newcommand{\Q}{\mathcal{Q}\xspace}

\newcommand{\QMClear}{\ensuremath{\Q_{clear^*}}\xspace}
\newcommand{\QOn}{\ensuremath{\Q_{on}}\xspace}
\newcommand{\QHanoi}{\ensuremath{\Q_{Hanoi}}\xspace}

\newcommand{\QTower}{\ensuremath{\Q_{tower}}\xspace}
\newcommand{\QBlocks}{\ensuremath{\Q_{blocks}}\xspace}

\newcommand{\hector}[1]{\textcolor{red}{Hector: #1}}

\title{On Policy Reuse: An Expressive Language for Representing and Executing General Policies that Call Other Policies}
\author{%
  Blai Bonet\textsuperscript{\rm 1},
  Dominik Drexler\textsuperscript{\rm 2},
  Hector Geffner\textsuperscript{\rm 3,2}
}
\affiliations{
  \textsuperscript{\rm 1}Universitat Pompeu Fabra, Spain \\
  \textsuperscript{\rm 2}Link\"oping University, Sweden \\
  \textsuperscript{\rm 3}RWTH Aachen University, Germany \\
  bonetblai@gmail.com, dominik.drexler@liu.se, hector.geffner@ml.rwth-aachen.de
}

\newcommand{\citeay}[1]{\citet{#1}}

\begin{document}
\allowdisplaybreaks

\maketitle

\begin{abstract}
    Recently, a simple but powerful language for expressing and learning general policies
    and problem decompositions (sketches) has been introduced in terms of 
    rules defined over  a set of Boolean and numerical  features.
    In this work, we consider three extensions of this language aimed at making
    policies and sketches more flexible and reusable: internal memory states, as in finite
    state controllers; indexical features, whose values are a function of the state
    and a number of internal registers that can be loaded with objects; and 
    modules that wrap up policies and sketches and allow them to call
    each other by passing parameters. 
%     4) an execution driver that implements a caller/callee protocol
%     and resolves the implicit non-determinism in the policies and
    %     sketches.
    In addition, unlike general policies that select state transitions
    rather than ground actions, the new language allows for the selection of
    such actions.
    %%In addition, unlike general policies  that   select actions indirectly
    %%by  selecting state transitions,     the new language allows for the
    %%selection of ground actions directly.
    The expressive power of the resulting language for %recombining
    policies and sketches is illustrated through a number of examples.
  % The problem of learning policies and sketches in the new language, from the
  %   bottom up, is  left for future work.
\end{abstract}

\section{Introduction}
%\vskip -.5em

%Classical planners solve problems over large state spaces by exploiting problem structure.
%Domain-independent methods assume, for example, that subgoals are independent when deriving
%heuristics \cite{bonet-geffner-aij2001,hoffmann-nebel-jair2001}, or that subproblems have low width
%\cite{lipovetzky-geffner-ecai2012,lipovetzky-geffner-aaai2017}. Domain-dependent methods, on the other
%hand,  make problem structure explicit in the form of hierarchies that express how tasks decompose into
%subtasks \cite{erol-et-al-aaai1994,georgievski-aiello-aij2015,bercher-et-al-ijcai2019}.

A new language for representing problem structure explicitly has
been introduced recently in the form of \emph{sketches} \cite{bonet-geffner-aaai2021,bonet:width2023}.
Sketches are collections of rules of the form $\xrule{C}{E}$ which are
defined over a set of Boolean and numerical domain features $\Phi$,
where $C$ expresses Boolean conditions on the features, and $E$ expresses qualitative changes in their values.
Each sketch rule captures a subproblem: the problem of going from a state $s$
whose feature values satisfy the condition $C$, to a state $s'$ where the
feature values change with respect to $s$ in agreement with $E$.
The language of sketches is powerful as it can encode everything from simple
goal serializations to full general policies.

The width of a sketch for a class of problems bounds the complexity of
solving the resulting subproblems \cite{bonet-geffner-aaai2021,bonet:width2023}.
For example, a sketch of width $k$ decomposes problems into subproblems that
are solved by the IW algorithm in time exponential in $k$ \cite{lipovetzky-geffner-ecai2012}.
Sketches provide a direct generalization of policies, which are sketches of
width $0$ that result in subproblems that can be solved in a single step.  % \cite{bonet:width2023}.

Combinatorial optimization methods for learning policies and sketches of this form has been  developed by
\citeay{frances-et-al-aaai2021} and \citeay{drexler-et-al-icaps2022}
where both the rules and the features  are obtained. % by solving a combinatorial optimization problem.

The current language of policies and  sketches, however, does not support the \emph{reuse}
of  other  policies and  sketches, and assumes instead  that the top goals are set up
externally.\footnote{For each predicate $p$ appearing in
  the goal, a new predicate $p_G$ is introduced with the same arity as $p$.
  The atom $p(c)$ in the state $s$ means that $p(c)$ is true in $s$, while
  $p_G(c)$ in $s$ means that $p(c)$ must be true in the goal.}
Yet for \emph{reusing}  policies and sketches,  goals must be set up internally
as well; e.g.,  for reusing a general policy $\pi_{on}$ that achieves the atom
$on(x,y)$ for any blocks $x$ and $y$ within a policy $\pi_T$ that builds a tower,
it is necessary for $\pi_T$ to repeatedly call $\pi_{on}$
on the right sequence of blocks $x$ and $y$.

In this work, we develop an extension of the language of policies and sketches
that accommodates reuse. %policy and sketch reuse.
In the resulting framework,  policies and sketches may call other policies and
sketches while passing them arguments.
For example, one can learn a policy $\pi_T$ for building towers of blocks
that uses a policy $\pi_{on}$ for putting one block on top of another, and
then reuse $\pi_T$ for learning a policy that builds any configuration of blocks.
%a policy  $\pi_{on}$ for putting one block on top of another, and use the learned policy
%for learning a policy $\pi_T$ for building a tower of blocks, and reuse this policy in turn
%for learning to build any configuration of blocks.

The language extensions involved are basically three:  \emph{memory states}, as in
finite state controllers for sequencing behaviors, \emph{indexical features},
whose values are a function of the state and a number of \emph{internal registers} that can be loaded with objects,
and \emph{modules,} that wrap policies and sketches so that they can be {reused} by other policies or sketches.
In this paper, we do not address the \emph{learning} problem, but the \emph{representation} problem,
as in order to learn these policies bottom up, one must be able to represent them.
%by passing them suitable parameters.
% For example, an existing policy $\pi_1$ for getting hold of the object in register \preg
% can be reused by a policy $\pi$ for delivering some package to the target cell.

The paper is structured as follows.
The next two sections cover related work and review planning
and sketches.
Then, two sections introduce extended policies and sketches,
and their semantics, and the following section introduces
modules, call rules, and action calls.
The last section contains a discussion with conclusions and
directions for future work.

\section{Related Work}
\label{sect:related-work}

\textbf{General policies.}
The paper builds on %a research thread that introduced the
notions of width,
generalized rule-based policies and sketches of bounded width, and %as well as
methods  for learning them \cite{lipovetzky-geffner-ecai2012,bonet-geffner-ijcai2018,bonet:width2023,frances-et-al-aaai2021,drexler-et-al-icaps2022,drexler-et-al-kr2023}. The problem of  representing and learning  general policies has a long history \cite{khardon-ai1999,martin-geffner-ai2004,fern-et-al-jair2006},
and general plans have also been represented  in logic \cite{srivastava-et-al-aij2011,illanes-mcilraith-aaai2019},
and  neural nets \cite{sid:sokoban,sylvie:asnet,bueno-et-al-aaai2019,rivlin-et-al-icaps2020wsprl,mausam:dl2,stahlberg-et-al-icaps2022,stahlberg-et-al-kr2022,stahlberg-et-al-kr2023}.
% by forms of first-order regression \cite{boutilier-et-al-ijcai2001,wang-et-al-jair2008,otterlo-rl2012,sanner-boutilier-ai2009},

% A limitation of these approaches is the lack of a language to accommodate
% policies that may call other policies by passing parameters;
% a limitation that is addressed in this work.

\smallskip
\noindent
\textbf{Planning programs and inductive programming}. Planning programs have been proposed as a  language for representing and learning
general  policies that make use of a number of programming language constructs \cite{javi1,javi2,javi3}. However, without a rich feature language
for talking about goals, states, and their relation, planning programs are basically limited to families of tasks where the goal is fixed;
like  ``putting all blocks on the table'',  but  not for ``building a given tower''.\footnote{
 The goal is fixed in a class $\Q$ of planning problems when for any $P$ in $\Q$, the goal $G$ of $P$ is  determined by its initial state.}
The problem of program reuse, which is related to the problem of policy reuse, has been considered in program synthesis and  inductive programming
\cite{inductive,dreamer}.

\smallskip
\noindent
\textbf{Deictic representations.} The use of registers to store objects  in the extended  language of policies and
sketches is closely related to the  use of indices and visual markers in deictic or
indexical representations  \cite{chapman:penguins,agre-chapman-ras1990,ballard-et-al-bbs1996}.
The computational value of such representations, however, has not been clear \cite{finney-et-al-arxiv2013}.
% , and more recent work has shown  that certain deictic representations in RL actually  harm performance
In our setting, indices (registers) make policies (and sketches) parametric and more expressive, as
(sub)policies can be reused by setting and resetting the values of registers as needed. %in a suitable way.

\smallskip
\noindent
\textbf{Hierarchical RL.} %  \alert{*** NOT CLEAR ***}
Hierarchical structures have been used in RL  in the form of options \cite{sutton-et-al-aij1999}, hierarchies of machines
\cite{parr1997reinforcement} and MaxQ hierarchies \cite{dietterich2000hierarchical},
and  a vast literature has explored  methods for learning hierarchical policies
\cite{mcgovern-barto-icml2001,machado-et-al-icml2017}, in some cases, 
representing goals and subgoals as part of an extended  state \cite{kulkarni:hrl,hierachies:pixels}.
In almost all cases, however, policies and subpolicies are learned jointly, not bottom up, and the information that is
conveyed by parameters, if they exist, is limited.

\Omit{
\smallskip
\noindent
\textbf{Intrinsic rewards and reward machines.}
Subgoal structure in RL has also been represented by means of
intrinsic rewards \cite{singh-et-al-ieeetamd2010,zheng-et-al-icml2020}
and reward machines \cite{icarte-et-al-jair2022,giuseppe:rewards}, which
are closely related to sketches. Three differences are that sketches are defined
in terms of state  features, not  additional variables, so there is no need for cross-products;
sketches have  a theory of width that tells us where problems
have to be split into subproblems; and finally, sketches have  a notion of termination
that ensures that subgoaling does not result in cycles \cite{bonet:width2023}.
}

\section{Planning Problems and Sketches}
\label{sect:planning-sketches}

A \textbf{planning problem}  refers to a  classical planning problem
$P=\tup{D,I}$ where $D$ is the planning domain and $I$ contains information about the
instance; namely, the objects in the instance, the initial situation, and the goal.
A class  $\Q$ of planning problems is a set of instances over a common domain $D$.
A \textbf{sketch} for a class of problems $\Q$ is a set of \textbf{rules} of the form \xrule{C}{E}
based on \textbf{features} over the domain $D$  which can be Boolean or numerical (i.e.\ non-negative integer valued)
\cite{bonet:width2023}. The condition $C$ is a conjunction of expressions like $p$ and $\neg p$ for
Boolean features $p$, and $\GT{n}$ and $\EQ{n}$ for numerical features $n$,
while  the effect $E$ is  a conjunction of expressions like $p$, $\neg p$, and $\UNK{p}$,
and $\DEC{n}$, $\INC{n}$ and $\UNK{n}$ for Boolean and numerical features $p$ and $n$,
respectively. A state pair $(s,s')$ over an instance $P$ in $\Q$ is \textbf{compatible} with
a rule $r$ if the state $s$ satisfies the condition $C$, and the change of  feature values from $s$ to $s'$ is
consistent with $E$. This is written as $s'\prec_r s$ and $s' \prec_R s$ if $R$ is a set  of rules that contains $r$.

A sketch $R$ for a class $\Q$ splits the problems $P$ in $\Q$
into \textbf{subproblems} $P[s]$ that are like $P$ but with initial
state $s$ (where $s$ is a reachable state in $P$), and goal states $s'$
that are either goal states of  $P$, or states $s'$ such that $s'\prec_R s$.
The  algorithm \siwR shown in Alg~\ref{Xalg:siwR} uses this problem decomposition
to solve problems $P$ in $\Q$ by solving subproblems $P[s]$
via the  IW algorithm \cite{lipovetzky-geffner-ecai2012}.
If the sketch has  bounded serialized width over  $\Q$ and is terminating,
\siwR  solves  any problem $P$ in $\Q$ in polynomial time
\cite{bonet:width2023,srivastava-et-al-aaai2011}.

\begin{algorithm}[t]
  \begin{algorithmic}[1]\small
    \smallskip
    \State \textbf{Input:} Sketch $R$ over features $\Phi$ that induces relation $\prec_R$
    \State \textbf{Input:} Planning problem $P$ with initial state $s_0$ on which the features in $\Phi$ are well defined
    %\smallskip
    \State $s\gets s_0$
    \State \textbf{while} $s$ is not a goal state of $P$ \textbf{do}
    \State\quad Run IW search from $s$ to find goal state $s'$ of $P$, or state
    \Statex\quad\quad $s'$ such that $s' \prec_R s$
    \State\quad \textbf{if} $s'$ is not found, \Return FAILURE
    \State\quad $s\gets s'$
    %\smallskip
    \State\Return path from $s_0$ to the goal state $s$
  \end{algorithmic}
  \caption{\siwR: sketch $R$ is used to decompose problem into subproblems,
    each solved with the IW algorithm.
  }
  \label{Xalg:siwR}
\end{algorithm}

\subsection{Serialized Width, Acyclicity, and Termination}

The \textbf{width} of a planning problem $P$ provides a complexity measure for finding an  optimal
plan for $P$. If $P$ has $N$ ground atoms and its width is bounded by $k$,
written  $w(P)\leq k$, an optimal plan for $P$ can be found in $\O(N^{2k-1})$ time
and $\O(N^k)$ space by running the algorithm \iw{k}, which is a simple breadth-first search
where newly generated states  are pruned if they do not make a tuple (set) of
$k$  or less atoms true for the first time in the search \cite{lipovetzky-geffner-ecai2012}.
If the width of $P$ is bounded but its value is unknown, %of the bound $k$ is not known,
a plan (not necessarily optimal) can be found by running the algorithm IW
with the same complexity bounds. IW calls \iw{i} iteratively with $i=0, \ldots,N$,
until $P$ is solved  \cite{lipovetzky-geffner-ecai2012}.
If $P$ has no solution, its width is defined as $w(P)\doteq\infty$, and if it has
a plan of length one, as  $w(P)\doteq0$. The width notion extends  to classes $\Q$
of problems: $w(\Q)\leq k$ iff $w(P)\leq k$ for each problem $P$ in $\Q$.
If $w(\Q)\leq k$ holds for class $\Q$, then any problem $P$ in $\Q$ can
be solved in \textbf{polynomial time} as $k$ is  independent of  the size of $P$.

A sketch $R$ has \textbf{serialized width} bounded by $k$ on a class $\Q$, denoted as
$w_R(\Q)\leq k$, if for any $P$ in $\Q$, the possible subproblems $P[s]$
have width bounded by $k$. % i.e., $w(P[s])\leq k$.
In such a case, every subproblem $P[s]$ that arises when  running the \siwR algorithm
on $P$ can be solved in \textbf{polynomial time.}

Finally, a  sketch $R$ is \textbf{acyclic} in $P$ if there
is no state sequence $s_0,s_1,\ldots,s_n$ in $P$ such that $s_{i+1}{\,\prec_R\,}s_i$, for $0\leq i<n$,
and $s_n=s_0$, and it is acyclic in $\Q$ if it is acyclic in each problem $P$ in $\Q$.
The \sieve algorithm \cite{srivastava-et-al-aaai2011} can check whether a sketch $R$ is
\textbf{terminating}, and hence acyclic, by just considering the rules in $R$
and the graph that they define. 
For a  terminating sketch $R$ with a bounded serialized width over  $\Q$,
\siwR is guaranteed to find a solution to any problem $P$ in $\Q$ in
\textbf{polynomial time} \cite{bonet:width2023}.

\medskip\noindent\textbf{Example 1.}
A sketch of width $0$ (i.e., a policy) for achieving the atom $on(x,y)$
for two blocks $x$ and $y$, on any Blocksworld instance, can be defined
with the numerical feature $n$ that counts the number of blocks above $x$,
$y$, or both, and the Boolean features \ON that represents whether the
goal holds (i.e., $x$ on $y$), $H$ that represents whether a block is
being held, and $H_x$ that represents whether the block $x$ is being
held.
The set of features is $\Phi=\{\ON,H,H_x,n\}$, and the rules are:

\medskip
\begin{tabular}{@{}l}
  $r_0\ =\ \prule{\neg\ON, \GT{n},\neg H,\neg H_x}{\DEC{n},H,\UNK{H_x}}$ \\
  %r_1\ =\ \prule{\neg\ON, \GT{n}, H}{\neg H,\neg H_x} \\
  $r_1\ =\ \prule{\neg\ON, \GT{n}, H}{\neg H,\neg H_x}$ \\
  $r_2\ =\ \prule{\neg\ON, \EQ{n}, \neg H, \neg H_x}{H, H_x}$ \\
  $r_3\ =\ \prule{\neg\ON, \EQ{n}, H, H_x}{\ON,\INC{n},\neg H, \neg H_x}$
\end{tabular}

\medskip
Rule $r_0$ says that in states where no block is being held and \GT{n}, 
picking a block above $x$ or $y$ is good, as it would decrement $n$ and
make $H$ true.
Rule $r_1$ says that in states where a block is being held and \GT{n},
putting the block away from $x$ or $y$ is good, as the resulting state
transition would not change the value of $n$. 
Last, rule $r_2$ says that in states where no block is held and \EQ{n},
picking up $x$ is good, and $r_3$, that in states where block $x$ is held
and \EQ{n}, putting $x$ on $y$ is good, as it results in \ON being true,
while also increasing $n$ and making $H$ and $H_x$ false.

A sketch of width 2 for \QOn is obtained instead by %not using $H_x$, and
replacing $r_0$ and $r_1$ with the rule \prule{\neg\ON,\GT{n}}{\DEC{n},\neg H},
and $r_2$ and $r_3$ with the rule \prule{\neg\ON,\EQ{n}}{\ON,\INC{n},\neg H}. \hfill\eex

\section{Extended Sketches}
\label{sect:ideas}

The first two extensions of  sketches are introduced next.

\subsection{Finite Memory}

The first language extension adds memory in the form of a finite number of
\emph{memory states} $m$:

\begin{definition}[Sketches with Memory]
  \label{def:sfm}
  A sketch with finite memory is a tuple $\tup{\M,\Phi,m_0,R}$ where $\M$
  is a finite set of memory states, $\Phi$ is a set of features, $m_0\in\M$ is
  the initial memory state, and $R$ is a set of rules extended with memory
  states. Such rules have  the form \xrule{(m,C)}{(E,m')} where \xrule{C}{E} is
  a standard sketch rule, and $m$ and $m'$ are memory states in $\M$.
\end{definition}

When the current memory state is $m$, only rules of the form \xrule{(m,C)}{(E,m')} apply.
If $s$ and $m$ are the current state and memory, respectively,
and $s'$ is a state reachable from $s$ such that the pair $(s,s')$
is compatible with the rule \xrule{C}{E},
then moving to state $s'$ and setting the memory to $m'$
is compatible with the extended rule \xrule{(m,C)}{(E,m')}.
%In certain cases, a rule like \xrule{(m,C)}{(true,m')},
%where \emph{true} is a feature that is true in all states,
%allows a transition in the memory state from  $m$ to $m'$ without
%involving a transition from a state $s$ into a different state $s'$.
%These rules are abbreviated as \xrule{(m,C)}{(\{\},m')}.
%\blai{** I think we should move this noop below, after introducing internal rules,
%and $m$ in this case should be internal to avoid the IW search to cope with these
%noops **}
In displayed listings, rules \xrule{(m,c)}{(E,m')} are
written as $m\,\|\,C\mapsto E\,\|\,m'$ to improve readability.

\medskip\noindent\textbf{Example 2.}
\citet{hanoi:aaai2023} describe a simple policy for solving the class
\QHanoi of Towers-of-Hanoi instances with 3 pegs, where a tower in the
first peg is to be moved to the third peg.
%The policy is expressed by referring to the relative size of the disk
%being moved, among the top disks at each peg, either a movement of the
%smallest disk, or a movement of the other disk:
%%%smallest disk (2 such possible movements), or a movement of the other disk (one such possible movement):
%\begin{quote}
In this policy, actions alternate between moving the smallest top disk
and moving the second smallest top disk.
When moving the smallest top disk, it is always moved to the ``left'',
towards the first peg, except when it is in the first peg that is then
moved to the third peg.
%If the smallest disk is on the first peg, it is moved to the third peg.
In the alternate step, the second smallest top disk is moved on top of
the third smallest top disk, as no other choice is available.
%When moving the second smallest top disk, in the alternate step,
%it must be moved on top of the third smallest top disk, as no other choice is available.
%\end{quote}

The movements can be expressed in the language of sketches with
three Boolean features $p_{i,j}$, $1 \leq i < j \leq 3$, that are true
if the top disk at peg $i$ is smaller than the top disk at peg $j$.
For example, the smallest disk is at the first (resp.\ third) peg iff
$p_{1,2}\land p_{1,3}$ (resp.\ $\neg p_{1,3} \land \neg p_{2,3}$) holds.

The alternation of actions is obtained with two memory states $m_0$ and
$m_1$ of which $m_0$ is the initial memory state. The rules implementing
the policy  are then:

\medskip
\begin{tabular}{@{}l}
  \textcolor{black}{\it\% Movements of the smallest disk}                                 \\
  $r_0\ =\ \nrule{m_0}{p_{1,2},p_{1,3}}{\UNK{p_{1,2}},\neg p_{1,3}, \neg p_{2,3}}{m_1}$      \\ %& Move smallest disk from peg 1 to peg 3 \\[2pt]
  $r_1\ =\ \nrule{m_0}{\neg p_{1,2},p_{2,3}}{p_{1,2},p_{1,3},\UNK{p_{2,3}}}{m_1}$            \\ %& Move smallest disk from peg 2 to peg 1 \\[2pt]
  $r_2\ =\ \nrule{m_0}{\neg p_{1,3}, \neg p_{2,3}}{\neg p_{1,2},\UNK{p_{1,3}},p_{2,3}}{m_1}$ \\[.8em] %& Move smallest disk from peg 3 to peg 2 \\[1em]

  \textcolor{black}{\text{\it\% Movements of the second smallest disk}}                   \\
  $r_3\ =\ \nrule{m_1}{p_{1,2},p_{1,3},p_{2,3}}{\neg p_{2,3}}{m_0}$                          \\ %& Move other disk from peg 2 to peg 3 \\[2pt]
  $r_4\ =\ \nrule{m_1}{p_{1,2},p_{1,3},\neg p_{2,3}}{p_{2,3}}{m_0}$                          \\ %& Move other disk from peg 3 to peg 2 \\[2pt]
  $r_5\ =\ \nrule{m_1}{\neg p_{1,2},p_{1,3},p_{2,3}}{\neg p_{1,3}}{m_0}$                     \\ %& Move other disk from peg 1 to peg 3 \\[2pt]
  $r_6\ =\ \nrule{m_1}{\neg p_{1,2},\neg p_{1,3},p_{2,3}}{p_{1,3}}{m_0}$                     \\ %& Move other disk from peg 3 to peg 1 \\[2pt]
  $r_7\ =\ \nrule{m_1}{p_{1,2},\neg p_{1,3},\neg p_{2,3}}{\neg p_{1,2}}{m_0}$                \\ %& Move other disk from peg 1 to peg 2 \\[2pt]
  $r_8\ =\ \nrule{m_1}{\neg p_{1,2},\neg p_{1,3},\neg p_{2,3}}{p_{1,2}}{m_0}$                   %& Move other disk from peg 2 to peg 1
\end{tabular}

\medskip
The policy corresponds to the sketch with memory \tup{M,\Phi,m_0,R}
where $M{\,=\,}\{m_0,m_1\}$, $\Phi{\,=\,}\{p_{i,j}:1\leq i<j\leq 3\}$, and
$R$ is the set of above rules. Notice that this policy is guaranteed
to solve only instances of Towers-of-Hanoi with 3 pegs where the initial
state contains a single tower in the first peg that is to be moved to
the third peg. \hfill\eex

\subsection{Registers, Indexicals, Concepts, and Roles} % Changed title to fit in one line

The  second  extension introduces internal memory in the form of \emph{registers}.
Registers store objects, which can be referred to in features
that become indexical or parametric, as their value changes
when the object in the register changes; e.g., the number of blocks
above the block in register zero.
The objects that can be placed into the registers $\Reg=\{\reg{0},\reg{1},\ldots\}$
are selected by two new classes of features called
\emph{concepts} and \emph{roles} that denote sets of objects
and set of object pairs, respectively \citep{baader-et-al-2003}.

Concept features or simply concepts are denoted in sans-serif font such as `\C',
and in a state $s$, they denote the set of objects in the problem $P$
that satisfy the unary predicate `\C'. % in $s$.
Role features or simply roles are  also denoted in sans-serif font  such as `\R',
and in a state $s$,  they denote the set of object pairs in the problem $P$  that satisfy the binary relation `\R'.
Concept and  role features are also used as numerical features, e.g., as  conditions `$\GT{\C}$' or effects `$\DEC{\R}$',
with the understanding that the corresponding numerical feature is given by the cardinality of the concept \C or role \R
in the state; namely, the number of objects or object pairs  in their denotation.

While a plain feature is a function of the problem state, an indexical or parametric feature
is  a function of the problem state and the value of the registers;
like ``the distance of the agent to the object stored in \reg{0}''.
When the value of a  register \preg changes, the denotation of
indexical features that depend on the value of the register may change as well.
We denote by $\Phi(\preg)$ the subset of features in $\Phi$ that refer to (i.e., depend on the value of)  register \preg.
The set of features $\Phi$ is assumed to contain a (parametric) concept for each register \preg, whose denotation is the singleton that contains
the object in \preg, and that is also denoted by \preg.

The extended sketch language provides \emph{load effects} of the form
\Load{\C}{\preg} for updating the value of registers for a concept
\C and register \preg; an expression that indicates that the content of the
register \preg is to be set to \emph{any object} in the (current)
denotation of \C, a choice that is \emph{non-deterministic}.
Loading an object into register \preg can be thought as placing
the \emph{marker} \preg on the object.
A rule with  effect \Load{\C}{\preg} has the condition $\GT{\C}$
to ensure that $\C$ contains some object. Likewise, since a load may change the denotation of features, the effect
of a load rule on register \preg is assumed to  contain also  the  extra effects $\UNK{\phi}$
for the features $\phi$ in $\Phi(\preg)$, and no other effects. Formally,

\begin{definition}[Extended Sketch Rules]
  \label{def:rules:ext}
  An \textbf{extended rule} over the features $\Phi$ and memory \M has the form \xrule{(m,C)}{(E,m')}
  where $m$ and $m'$ are memory states,  $C$ is a Boolean condition on the features,
  and either $E$ expresses changes in the feature values, as usual, or
%
%   $C$ is a set of conditions of the form $p$, $\neg p$, $\EQ{n}$, and
%   $\GT{n}$ for Boolean and numerical features $p$ and $n$ in $\Phi$, and
%  $E$ is a set of effects of the form $p$, $\neg p$, $\UNK{p}$ for Boolean $p$,
%  $\DEC{n}$, $\INC{n}$, and $\UNK{n}$ for numerical $n$. In addition,
   $E$ contains one load  effect of the form \Load{\C}{\preg} for some concept \C and register \preg.
  %If $E$ contains $\DEC{n}$, it may also contain $\EQ{n}$ or $\GT{n}$,
  If the latter case, $E$ must also contain uncertain effects $\UNK{\phi}$
  on  the features $\phi$ in  $\Phi(\preg)$, but no other effects.
  %In such a case, the memory $m$ must be internal, and the rule is called
  %a load or internal rule. If $m$ is external memory, $r$ is called an external rule.
\end{definition}

Rules with a load effect are called \emph{internal rules}, as they capture
changes in the internal memory, while the other rules are called \emph{external.}
For simplicity, it is assumed, that each load rule contains a single load effect,
and that internal and external rules apply in different memory states,
from now called \emph{internal}  and \emph{external} memory states.
%if a load  rule is applicable in a memory state,
%there is no other rule, internal or external, that does.
For convenience,  rules of the form \xrule{(m,C)}{(true,m')}, abbreviated as \xrule{(m,C)}{(\{\},m')},
are allowed to enable a change from memory state $m$ to memory state $m'$
under the  condition $C$.  Extended sketches are comprised of extended sketch rules:

\begin{definition}[Extended Sketch]
  \label{def:esketch}
  An \textbf{extended sketch} is a tuple \tup{\M,\Reg,\Phi,m_0,R} where \M and
  \Reg are finite sets of memory states and registers, respectively, $\Phi$ is
  a set of features, $m_0{\,\in\,}\M$ is the initial memory state, and $R$ is
  a set of extended $\Phi$-rules over memory \M.
  %   An extended sketch is well defined on a class $\Q$ if its set of features
  %   $\Phi$ is well defined on $\Q$.  Really needed?
  If $m$ is a memory state, $R(m)$ denote the subset of rules in $R$ of form
  \xrule{(m,C)}{(E,m')}.
\end{definition}

Extended sketches are also called indexical sketches, and
those of width 0 are called extended or indexical policies.

\medskip\noindent\textbf{Example 3.}
A different, indexical, policy $\pion^*$ for the class \QOn of
Blocksworld problems considered in Example~1, where the goal is
to achieve the atom $on(x,y)$, is expressed as the sketch
\tup{M,\R,\Phi,m_0,R} where $M{\,=\,}\{m_i{\,:\,}0{\,\leq\,}i{\,\leq\,}8\}$ has
9 memory states, and $\R{\,=\,}\{\reg{0},\reg{1}\}$ has two registers.
The set of features is $\Phi{\,=\,}\{\ON,H,H_x,A,\concept{N},\concept{T}_0,\concept{T}_1\}$
where $\ON$, $H$ and $H_x$ are as in Example~1, $A$ is true if the block in
\reg{1} is above $x$ or $y$, \concept{N} is the subset of blocks in $\{x,y\}$
that are \emph{not clear,} and the concept $\concept{T}_0$ (resp.\ $\concept{T}_1$)
contains the block, if any, that is directly above the block in \reg{0}
(resp.\ \reg{1}).
The set $R$ has the following 14 rules:

\medskip
\resizebox{.99\linewidth}{!}{
\begin{tabular}{@{}l}
  \textcolor{black}{\it\% Internal rules (update registers and internal memory)} \\
  \textcolor{black}{\small\it\% Initial border case when holding some block: apply rule $r_8$} \\
    $r_0\ =\ \nrule{m_0}{H}{}{m_4}$ \\[.5em]

  \textcolor{black}{\small\it\% Blocks $x$ and $y$ already clear: apply rule $r_{12}$} \\
     $r_1\ =\ \nrule{m_0}{\neg H,\EQ{\concept{N}}}{}{m_7}$ \\[.5em]

  \textcolor{black}{\small\it\% Place \reg{0} on some block in \concept{N}, and \reg{1} on topmost above \reg{0}} \\
     $r_2\ =\ \nrule{m_0}{\neg H,\GT{\concept{N}}}{\Load{\concept{N}}{\reg{0}},\UNK{\concept{T}_0}}{m_1}$ \\
     $r_3\ =\ \nrule{m_1}{\GT{\concept{T}_0}}{\Load{\concept{T}_0}{\reg{1}},\UNK{\concept{T}_1},\UNK{A}}{m_2}$ \\
     $r_4\ =\ \nrule{m_2}{\GT{\concept{T}_1}}{\Load{\concept{T}_1}{\reg{1}},\UNK{\concept{T}_1},\UNK{A}}{m_2}$ \\
     $r_5\ =\ \nrule{m_2}{\EQ{\concept{T}_1}}{}{m_5}$ \\[.5em] %\hfill\textcolor{black}{\small\it(go to external rule)} \\[.5em]

  \textcolor{black}{\small\it\% Iterate over next block above \reg{0}, or next block in \concept{N}} \\
     $r_6\ =\ \nrule{m_3}{\GT{\concept{T}_0}}{}{m_1}$ \\       %\hfill\textcolor{black}{\small\it(go to internal rule)} \\
     $r_7\ =\ \nrule{m_3}{\EQ{\concept{T}_0}}{}{m_0}$ \\[.8em] %\hfill\textcolor{black}{\small\it(go to internal rule)} \\[.5em]

  \textcolor{black}{\it\% External rules (involve state transitions)} \\
  \textcolor{black}{\small\it\% Handle initial border case: drop block being held} \\
     $r_8\ =\ \nrule{m_4}{H}{\neg H,\neg H_x,\UNK{\ON},\UNK{A},\UNK{\concept{N}},\UNK{\concept{T}_0},\UNK{\concept{T}_1}}{m_0}$ \\[.5em]

  \textcolor{black}{\small\it\% Pick block \reg{1} (topmost above \reg{0}) and put it away} \\
     $r_9\ =\ \nrule{m_5}{\neg H,A}{H,\neg A,\UNK{\concept{N}},\UNK{\concept{T}_0}}{m_6}$ \\
  $r_{10}\ =\ \nrule{m_6}{     H,\neg A}{\neg H}{m_3}$ \\ %\hfill\textcolor{black}{\small\it(go to internal rule)} \\
  $r_{11}\ =\ \nrule{m_6}{     H,\neg A}{\neg H,\DEC{\concept{N}}}{m_3}$ \\[.5em] %\hfill\textcolor{black}{\small\it(go to internal rule)} \\[.5em]

  \textcolor{black}{\small\it\% Pick up block $x$ and place on block $y$} \\
  $r_{12}\ =\ \nrule{m_7}{     \neg H_x}{H, H_x,\UNK{\concept{N}}}{m_8}$ \\
  $r_{13}\ =\ \nrule{m_8}{     H_x,\neg\ON}{\neg H,\neg H_x,\ON}{m_8}$
\end{tabular}
}

\medskip
The policy $\pion^*$ is \emph{indexical} \cite{agre-chapman-ras1990,ballard-et-al-bbs1996},
as it puts the \emph{``mark''} \reg{0} on a block in \concept{N} (rule $r_2$), while
a second \emph{mark} \reg{1} is moved up from the block on \reg{0} to the topmost block
above \reg{0} (rules $r_3$--$r_5$).
The block marked as \reg{1} is then picked and put away (rules $r_9$--$r_{11}$),
and the process repeats (loop on rules $r_6,r_3,r_4^+,r_5,r_9,r_{10},r_6$) until
the block in \reg{0} becomes clear. Then, another block in \concept{N}, if any,
is loaded into \reg{0} (rules $r_7,r_2$).
Once $x$ and $y$ become clear (i.e., $\EQ{\concept{N}}$), the block $x$ is
picked and placed on $y$ (rules $r_1$, $r_{12}$ and $r_{13}$).
The rule $r_0$ handles the special case when initially some block is being held.
%When the block in \reg{0} becomes clear, the process
%repeats until the counter \concept{N} becomes zero.
%When \concept{N} is zero, both $x$ and $y$ are clear.
%The policy then picks $x$ and puts it on top of $y$.
%The new policy, unlike the  old policy, does not  use of features for counting  blocks.
%The new  features are indeed  simpler than the old ones, as they make use of
%``marks'' (i.e.,  registers that  store or mark specific objects)
%to \emph{fix the attention} on specific blocks, something
%which non-indexical features cannot do.

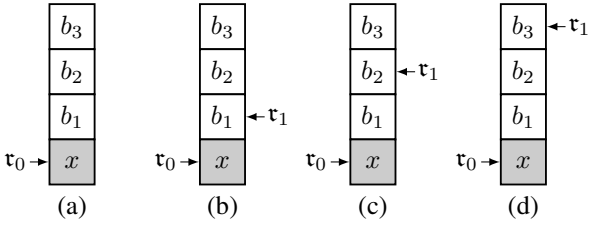
\begin{figure}
  \centering
  %\resizebox{\linewidth}{!}{
    \begin{tikzpicture}[]
      \node[draw, black, sharp corners, fill=white, line width=0.7pt, rectangle, minimum size=0.6cm]     (a-b3) at (0, 1.8) {$b_3$};
      \node[draw, black, sharp corners, fill=white, line width=0.7pt, rectangle, minimum size=0.6cm]     (a-b2) at (0, 1.2) {$b_2$};
      \node[draw, black, sharp corners, fill=white, line width=0.7pt, rectangle, minimum size=0.6cm]     (a-b1) at (0, 0.6) {$b_1$};
      \node[draw, black, sharp corners, fill=lightgrey, line width=0.7pt, rectangle, minimum size=0.6cm]  (a-x) at (0, 0)   {$x$};
      \node[align=left, inner sep=1pt] (a-label) at (0,-0.6) {(a)};
      \node[align=left, inner sep=1pt] (a-marker1) at (-0.75,0) {\reg{0}};
      \draw[-latex] (a-marker1) -- (a-x);

      \node[draw, black, sharp corners, fill=white, line width=0.7pt, rectangle, minimum size=0.6cm]     (b-b3) at (2, 1.8) {$b_3$};
      \node[draw, black, sharp corners, fill=white, line width=0.7pt, rectangle, minimum size=0.6cm]     (b-b2) at (2, 1.2) {$b_2$};
      \node[draw, black, sharp corners, fill=white, line width=0.7pt, rectangle, minimum size=0.6cm]     (b-b1) at (2, 0.6) {$b_1$};
      \node[draw, black, sharp corners, fill=lightgrey, line width=0.7pt, rectangle, minimum size=0.6cm]  (b-x) at (2, 0)   {$x$};
      \node[align=left, inner sep=1pt] (b-label) at (2,-0.6) {(b)};
      \node[align=left, inner sep=1pt] (b-marker1) at (2-0.75,0) {\reg{0}};
      \node[align=left, inner sep=1pt] (b-marker2) at (2.75,0.6) {\reg{1}};
      \draw[-latex] (b-marker1) -- (b-x);
      \draw[-latex] (b-marker2) -- (b-b1);

      \node[draw, black, sharp corners, fill=white, line width=0.7pt, rectangle, minimum size=0.6cm]     (c-b3) at (4, 1.8) {$b_3$};
      \node[draw, black, sharp corners, fill=white, line width=0.7pt, rectangle, minimum size=0.6cm]     (c-b2) at (4, 1.2) {$b_2$};
      \node[draw, black, sharp corners, fill=white, line width=0.7pt, rectangle, minimum size=0.6cm]     (c-b1) at (4, 0.6) {$b_1$};
      \node[draw, black, sharp corners, fill=lightgrey, line width=0.7pt, rectangle, minimum size=0.6cm]  (c-x) at (4, 0)   {$x$};
      \node[align=left, inner sep=1pt] (c-label) at (4,-0.6) {(c)};
      \node[align=left, inner sep=1pt] (c-marker1) at (4-0.75,0) {\reg{0}};
      \node[align=left, inner sep=1pt] (c-marker3) at (4.75,1.2) {\reg{1}};
      \draw[-latex] (c-marker1) -- (c-x);
      \draw[-latex] (c-marker3) -- (c-b2);

      \node[draw, black, sharp corners, fill=white, line width=0.7pt, rectangle, minimum size=0.6cm]     (d-b3) at (6, 1.8) {$b_3$};
      \node[draw, black, sharp corners, fill=white, line width=0.7pt, rectangle, minimum size=0.6cm]     (d-b2) at (6, 1.2) {$b_2$};
      \node[draw, black, sharp corners, fill=white, line width=0.7pt, rectangle, minimum size=0.6cm]     (d-b1) at (6, 0.6) {$b_1$};
      \node[draw, black, sharp corners, fill=lightgrey, line width=0.7pt, rectangle, minimum size=0.6cm]  (d-x) at (6, 0)   {$x$};
      \node[align=left, inner sep=1pt] (d-label) at (6,-0.6) {(d)};
      \node[align=left, inner sep=1pt] (d-marker1) at (6-0.75,0) {\reg{0}};
      \node[align=left, inner sep=1pt] (d-marker4) at (6.75,1.8) {\reg{1}};
      \draw[-latex] (d-marker1) -- (d-x);
      \draw[-latex] (d-marker4) -- (d-b3);
    \end{tikzpicture}
  %}
  \caption{Policy $\pion^*$ places markers \reg{0} and \reg{1} on blocks $x$ and $b_3$, respectively, for an example tower with 4 blocks.
    (a) The rule $r_2$ puts the marker \reg{0} on the block $x{\,\in\,}\concept{N}$; side effect is $\concept{T}_0{\,=\,}\{b_1\}$.
    (b) Rule $r_3$ initializes marker \reg{1} on block $b_1{\,\in\,}\concept{T}_0$; side effect is $\concept{T}_1{\,=\,}\{b_2\}$.
    (c) Rule $r_4$ moves marker \reg{1} one step above to block $b_2{\,\in\,}\concept{T}_1$; side effect is $\concept{T}_1{\,=\,}\{b_3\}$.
    (d) Another application of $r_4$ moves \reg{1} on block $b_3{\,\in\,}\concept{T}_1$; side effect is $\concept{T}_1{\,=\,}\emptyset$,
    and $r_5$ is the next rule to apply.
  }
  \label{fig:indexical}
\end{figure}

The new policy, unlike the one in Example~1, does not use features for
counting blocks. The features used are indeed computationally simpler as
they refer to ``markers'' (i.e., registers that store or mark specific
objects) that \emph{fix the attention} on specific blocks, something
which non-indexical features cannot do.
Figure~\ref{fig:indexical} illustrates the use of markers on a tower of
4 blocks where $x$ is at the bottom. \hfill\eex

\section{Formal Semantics and Termination}% of Extended Sketches}
\label{sect:formal}

Extended sketches are evaluated on planning states augmented with memory
states and register values.

\begin{definition}[Augmented states]
  \label{def:augmented}
  An \textbf{augmented state} for a  problem $P$ given an extended sketch  \tup{\M,\Reg,\Phi,m_0,R}
  is a tuple $\bar s=(s,m,\v)$ where $s$ is a  reachable state in $P$, $m$ is a memory state in \M, and \v is a vector of objects
  in $\Obj(P)^{\Reg}$ that tells the content of each register \preg, denoted   as $\v[\preg]$.
\end{definition}

Features $\phi$ are evaluated over pairs $(s,\v)$ made up of a state $s$ and a value
$\v$ for the registers. The value for $\phi$ at such a pair is denoted by $\phi(s,\v)$.

\begin{definition}[Compatible pairs]
  \label{def:satisfies}
  Let $r$ be an \textbf{external} $\Phi$-rule \xrule{(m,C)}{(E,m')}, and let \v
  be a valuation for the registers.
  A state $s$ satisfies the condition $C$ \textbf{given \v} if the feature
  conditions in $C$ are all true in $(s,\v)$.
  A state pair $(s,s')$ satisfies the effect $E$ \textbf{given \v} if the
  values for the features in $\Phi$ change from $s$ to $s'$ according to $E$;
  i.e., the following holds where $p$ is a Boolean feature, $n$ is a numerical,
  concept or role feature, and $\phi$ is any type of feature,
  \begin{enumerate}[1.]\itemsep0pt\topsep0pt
    \item if $p$ (resp.\ $\neg p$) is in $E$, $p(s',\v)=1$ (resp.\ $p(s',\v)=0$),
    \item if $\DEC{n}$ (resp.\ $\INC{n}$) is in $E$, $n(s,\v){\,>\,}n(s',\v)$ (resp.\ $n(s,\v){\,<\,}n(s',\v)$), and
    \item if $\phi$ is not mentioned in $E$, $\phi(s,\v)=\phi(s',\v)$.
    %\item if $\EQ{n}$ (resp.\ $\GT{n}$) is in $E$, then $n(s',\v)=0$ (resp.\ $n(s',\v)>0$).
  \end{enumerate}
  The pair $(s,s')$ is \textbf{compatible} with an external rule $r=\xrule{(m,C)}{(E,m')}$ \textbf{given \v},
  denoted as $s' \prec_{r/\v} s$, if given \v, $s$ satisfies $C$ and the pair satisfies $E$.
  The pair is compatible with a set of rules $R$ given \v, denoted as $s'\prec_{R/\v} s$,
  if it is compatible with some external rule in $R$ given \v.
\end{definition}

The search algorithm for extended sketches, called \siwRx and shown
in Alg.~\ref{Xalg:siwRx}, maintains the current memory state and register values,
implements the internal rules, and performs  IW searches to solve the subproblems
that arise when the memory state are external.

\begin{definition}[Subproblems]
  \label{def:subproblems}
  If $\bar s=(s,m,\v)$ is an augmented state for $P$, where $m$ is external memory, the subproblem $P[\bar s]$
  induced by $\bar s$ in $P$ is a planning problem which is like $P$ but with initial state $s$, and goal
  states that are either goal states of $P$, or states $s'$ such that $s' \prec_{r/\v} s$ for some rule $r$ in $R(m)$.
\end{definition}

Internal rules do not generate classical subproblems and do not change the planning state $s$
but they affect the  memory state and the register values, and with that, the definition of the
subproblems that follow. The  notion of \textbf{reduction}  captures how internal rules are
processed:

\begin{definition}[Reduction]
  \label{def:reduction}
  Let \tup{\M,\Reg,\Phi,m_0,R} be an extended sketch for a planning
  problem $P$, and let $(s,m,\v)$ be an augmented state for $P$ where
  $m$ is an internal memory state in \M.
  The pair \tup{(s,m,\v),(s,m',\v')} is a \textbf{reduction step} if
  there is a  rule $r$ in $R$ of form \xrule{(m,C)}{(E,m')} such that
  1)~$s$ satisfies the condition $C$ given \v, and
  2)~either $E$ is a dummy true effect and
  $\v'=\v$, or $E$ contains \Load{\C}{\preg}, and $\v'[\preg]\in\C(s,\v)$.
%  2)~
%  $\v'$ is equal to $\v$, except if \Load{\C}{\preg} is in $E$,
%  in which case $\v'[\preg]\in\C(s,\v)$.
  A sequence of reduction steps starting at $(s,m,\v)$ and ending at $(s,m',\v')$
  where $m'$ is external is called a \textbf{reduction}, and it is denoted
  by \Reduct{(s,m,\v)}{(s,m',\v')}.
  For convenience,  if $m$ is external, the triplet $(s,m,\v)$ is
  assumed to reduce to itself, written \Reduct{(s,m,\v)}{(s,m,\v)}.
  %\hector{Is this last sentence needed? If so, make it passive}
  %\blai{Yes, right below to define initial augmented state when $m_0$ is external.
  %  What do you mean by passive?}
\end{definition}

An \textbf{initial augmented state} for problem $P$ is of the form $(s_0,m,\v)$
where $s_0$ is the initial state in $P$, and \Reduct{(s_0,m_0,\v_0)}{(s_0,m,\v)}
for the initial memory $m_0$ and some $\v_0$ in $\Obj(P)^\Reg$.
There may be different initial augmented states for $P$ that differ
in their memory and/or the contents of the registers.
Each such initial augmented state defines an \textbf{initial subproblem}
$P[s_0,m,\v]$ (cf.\ Definition~\ref{def:subproblems}).

\begin{definition}[Induced subproblems]
  \label{def:induced}
  Let \tup{\M,\Reg,\Phi,m_0,R} be an extended sketch for a planning
  problem $P$ with initial state $s_0$.
  Let us consider a subproblem $P[\bar s]$ for $\bar s=(s,m,\v)$ where $m$ is external, and let $s'$
  be a state reachable from $s$ with $s'\prec_{r/\v} s$ for
  some rule $r$ in $R(m)$.
  Then,
  \begin{enumerate}[1.]\itemsep0pt\topsep0pt
    \item subproblem $P[s',m',\v]$ is \textbf{induced} by  subproblem $P[\bar s]$   if $r=\xrule{(m,C)}{(E,m')}$ and $m'$ is external memory.

    \item Subproblem $P[s',m'',\v']$ is \textbf{induced} by  subproblem $P[\bar s]$ if $r\,{=}\,(m,C)\,{\mapsto}\,(E,m')$,
      $m'$ is internal memory, and \Reduct{(s',m',\v)}{(s',m'',\v')}.

  \end{enumerate}
  The collection \closure{P} of induced subproblems is the {smallest} set such that
  1)~\closure{P} contains all the initial subproblems, and
  2)~$P[s',m',\v']$ is in \closure{P} if  $P[s,m,\v]$  is  in \closure{P} and the first subproblem is induced by the second.
\end{definition}

In this definition, subproblems $P[s,m,v]$ where $m$ is internal are ``jumped over''
so that the subproblems that make it into  \closure{P} all have memory states $m$ that are external, and hence
represent classical planning problems and not bookkeeping operations. The extended sketch $R$ is said to be \textbf{reducible} in $P$
if for any reachable augmented state $(s,m,\v)$ in $P$ where $m$ is an internal state,
there is an augmented state  $(s,m',\v')$ such that \Reduct{(s,m,\v)}{(s,m',\v')}.
A non-reducible sketch is one in which the executions can cycle or  get stuck
while performing internal memory operations.

\begin{definition}[Sketch width]
  The \textbf{width} of a reducible sketch  $R$ over a planning problem $P$ is bounded by non-negative
  integer $k$, denoted by $w_R(P)\leq k$, if the width of each subproblem in
  \closure{P} is bounded by $k$.
  The \textbf{width}  of a reducible sketch $R$ over a class of problems $\Q$ is bounded by $k$,
  denoted by $w_R(\Q)\leq k$, if $w_R(P)\leq k$ for each $P$ in $\Q$.
\end{definition}

The \textbf{width} is zero for the indexical sketch $\pion^*$ in Example~3, % \ref{ex:policy:on:indexical},
as the sketch represents a policy where each subproblem is solved in a single step.

\subsection{Termination for Extended Sketches}

Termination is a key property for sketches that guarantees
acyclicity, and that only a polynomial number of subproblems
may appear when solving \emph{any} problem $P$ where the
features are well defined \cite{bonet:width2023}.
Termination can be tested in polynomial time
by a suitable adaptation of the \sieve algorithm
\cite{srivastava-et-al-aaai2011,bonet:width2023}.
If the extended sketch is reducible,  terminating, and has bounded serialized width over a
class $\Q$,  any problem $P$ in $\Q$ can be solved in polynomial time with
the procedure \siwRx  shown in Alg.~\ref{Xalg:siwRx}.
%, as each subproblem
%can be solved in polynomial time, and their number is polynomially bounded.

\begin{algorithm}[t]
    \begin{algorithmic}[1]\small
      \smallskip
      \State \textbf{Input:} Extended sketch \tup{\M,\Reg,\Phi,m_0,R} %that defines $\prec_{R/\v}$
      \State \textbf{Input:} Planning problem $P$ with initial state $s_0$ on which the features in $\Phi$ are well defined
      \State $\bar s \gets (s_0,m_0,\v)$ for some $\v\in\Obj(P)^\Reg$
      \State \textbf{while} $s$ in $\bar s=(s,m,\v)$ is not a goal state of $P$ %\textbf{do}
      \State\quad\textbf{if} $m$ is internal memory %\textbf{then}
      \State\quad\quad Find rule $r=\xrule{(m,C)}{(E,m')}$ with $s,\v\vDash C$
      \State\quad\quad\textbf{if} $r$ is not found, \Return FAILURE \hfill\textcolor{black}{\% Irreducible}
      \State\quad\quad\textbf{if} $\Load{\C}{\preg}$ in $E$, $\v[\preg]\gets o$ for some $o\in\C(s,\v)$
      \State\quad\quad $\bar s\gets(s,m',\v)$
      \State\quad\textbf{else} \hfill\textcolor{black}{\% $m$ is external, solve subproblem}
      \State\quad\quad Run IW search from $s$ to find goal state $s'$ of $P$, or
      \Statex\quad\quad\quad state $s'$ such that $s'\prec_{r/\v} s$ for some (external)
      \Statex\quad\quad\quad rule $r=\xrule{(m,C)}{(E,m')}$ in $R$
      \State\quad\quad\textbf{if} no such state is found, \Return FAILURE %\par
          %\hskip\algorithmicindent \Comment{\textcolor{black}{(Subproblem $P[s,m,\v]$ is unsolvable)}}
          \State\quad\quad $\bar s\gets(s',m',\v)$
        %\EndIf
      %\EndWhile
      \State\Return path from $s_0$ to the goal state $s$
    \end{algorithmic}
  \caption{\siwRx uses the extended sketch $R$ to decompose problem $P$
    into subproblems that are solved with IW.
    Completeness of \siwRx is captured in Theorem~\ref{thm:main}.
  }
  \label{Xalg:siwRx}
\end{algorithm}

\begin{theorem}[Termination]
  \label{thm:main}
  If  the extended  sketch $R$ is \textbf{reducible}, \textbf{terminating}, and has a  \textbf{serialized width}
  bounded by $k$ over the class of problems  $\Q$, then \siwRx finds plans for any problem $P$ in $\Q$ in \textbf{polynomial time.}
\end{theorem}

\Omit{% SPACE REASONS
  \begin{proof}[Proof (main ideas, no details):]
    Let us suppose that \sieve accepts the sketch. It does so by eliminating
    all the cycles that have edges induced by the internal rules, by removing
    at least one edge $e$ in each such cycle. The edge $e$ cannot be labeled
    with an effect $\DEC{n}$ for a parametric feature defined on a register
    \preg that appears in a \Load{\U}{\preg} effect in the cycle (because
    the cycle then contains an edge with $\UNK{n}$ in its label).
    Therefore, the cycle is eliminated by a feature that decreases infinitely
    often, and never increases.
    Once all cycles involving internal rules are eliminated, the remaining
    of the graph is similar to the graph for a standard sketch, which is
    properly handled by \sieve. Hence, \siwRx terminated on any problem $P$
    on which the features in $\Phi$ are well defined. The termination cannot
    be due to IW failing to find the state $s'$ as the serialized width on
    $P$ is bounded, and it cannot be due to failing to reduce internal
    memory to external memory as the sketch is regular.
    The complexity bounds follow from the bounded serialized width assumption.
  \end{proof}
}

%\section{Wrapping Policies and Sketches into Modules for Reuse}

\section{Reusable Modules}
\label{sect:modules}

Policies and  sketches are  wrapped into \emph{modules} to enable calls among them.
A module is a named tuple \tup{\args,Z,M,\Reg,\Phi,m_0,R} where
$\args=\tup{x_1,x_2,\ldots,x_n}$ is a tuple of arguments, each one
being either a \emph{static concept or role}, $Z$
and $\Phi$ are sets of features, $M$ is a set of memory
states, \Reg is a set of registers, $m_0\in M$ is the initial memory
state, and $R$ is a set of rules.
The features in $\Phi$ are the ones mentioned in the
sketch rules in $R$, and their denotation may depend on the value
of the arguments and the features in $Z$, and the contents of the registers.
The sketch  $R$ in a module may also contain two new types of \emph{external} rules, \emph{call rules} and \emph{do rules}.
The first type permits to call other modules; the second type permits the \emph{direct execution of  ground actions of  the planning problem}.
The value for the arguments in either case is given by the features in $\Phi$ or $Z$, and the arguments of the module;
the difference between $\Phi$ and $Z$ is that the rules in $R$ must track the
changes for the features in $\Phi$, while $Z$ can be used to define features
in $\Phi$, appear in conditions of rules, and provide values to arguments in call and do rules.
If the name of the module is \module{name}, a \emph{call} to the module is expressed as $\module{name}(x_1,x_2,\ldots,x_n)$
where $x_1, \ldots, x_n$ are the module arguments.

Call and do \emph{rules} are of the form \call{m}{C}{\module{name}}{v_1,v_2,\ldots,v_n}{m'}
where $m$ is internal memory, $C$ is a condition, \module{name} is a module
or action schema name, and each value $v_i$ is of an appropriate type:
% BLAI: Fixed: call/do rules are *external* rules. This can be checked in the algorithm.
%\footnote{\hector{Check
%  if this affects definition of reduction. Before, we said that internal rules are reduced.
%  Actually, we said that internal rules with $E$ true or $E$ with a load action are reduced.
%  So looks that this is ok, and it's a different $E$, yet ..}
%  \blai{We don't provide any theoretical claim for modules so no need to check this.
%  Indeed, in the current setting, nothing can be guaranteed as cyclic behaviour is possible
%  and not testable}
%}
for \emph{do rules},  $v_i$ must be a concept, while for \emph{call rules},
$v_i$ can be either a concept or a role. The idea is that if a \emph{call rule} is used, the sketch associated
with the module \module{name} is executed until no  rules are applicable,
and the control is then returned back to the caller at the memory state $m'$.
For \emph{do rules}, an applicable ground action of the form
$\module{name}(o_1,o_2,\ldots,o_n)$, where the object $o_i$ belongs to
the denotation of concept $v_i$, must be applied at the current state to
make a transition to a successor state, and control is then returned
back to the caller at the memory state $m'$.

The execution model for handling modules involves a stack as described below.
Modules call each other by passing arguments but do not get back any values.
The ``side effects'' of a module are in the problem state $s$ that must be driven
eventually to a goal state.

\medskip\noindent\textbf{Example 4.}
Module $\module{on}(\X,\Y)$ for a policy for the class \QOn that implements
a variant of the policy $\pion^*$ in Example~3.
Indeed, \emph{by directly executing ground actions}, the policy uses simpler
features and can be executed model-free: there is no need to generate the
possible successors for choosing the ground action to do.
This is directly encoded in the policy.
For example, the indexical Boolean feature $A$ is no longer needed because
the block being held is put away on the table with a \module{Putdown} action.

The module $\module{on}(\X,\Y)$ has as parameters the concepts \X and \Y
that are assumed to be singletons containing the blocks $x$ and $y$,
respectively.
The module uses \emph{do rules} to avoid references to complex concepts;
it only uses the concepts \concept{B} for all blocks, \concept{N} for
the blocks in $\X\cup\Y$ that are not clear, and the indexical concept
$\concept{T}_0$ (resp.\ $\concept{T}_1$) for the block directly above \reg{0} (resp.\ \reg{1}), if any.
The other feature is the Boolean $T_{\concept{x}}$ that is true iff \X is on the table.
The module uses 9 memory states, two registers, and the following rules:

\medskip
\begin{tabular}{@{}l}
  \textcolor{black}{\it\% Internal rules (update registers and internal memory)} \\
     $r_0\ =\ \nrule{m_0}{H}{}{m_4}$ \\
     $r_1\ =\ \nrule{m_0}{\neg H,\EQ{\concept{N}}}{}{m_7}$ \\
     $r_2\ =\ \nrule{m_0}{\neg H,\GT{\concept{N}}}{\Load{\concept{N}}{\reg{0}},\UNK{\concept{T}_0}}{m_1}$ \\
     $r_3\ =\ \nrule{m_1}{\GT{\concept{T}_0}}{\Load{\concept{T}_0}{\reg{1}},\UNK{\concept{T}_1}}{m_2}$ \\
     $r_4\ =\ \nrule{m_2}{\GT{\concept{T}_1}}{\Load{\concept{T}_1}{\reg{1}},\UNK{\concept{T}_1}}{m_2}$ \\
     $r_5\ =\ \nrule{m_2}{\EQ{\concept{T}_1}}{}{m_5}$ \\
     $r_6\ =\ \nrule{m_3}{\GT{\concept{T}_0}}{}{m_1}$ \\
     $r_7\ =\ \nrule{m_3}{\EQ{\concept{T}_0}}{}{m_0}$ \\[.8em]
  \textcolor{black}{\it\% External rules (involve state transitions)} \\
     $r_8\ =\ \dcall{m_4}{}{\module{Putdown}}{\concept{B}}{m_0}$ \\
     $r_9\ =\ \dcall{m_5}{}{\module{Unstack}}{\reg{1},\concept{B}}{m_6}$ \\
  $r_{10}\ =\ \dcall{m_6}{}{\module{Putdown}}{\reg{1}}{m_3}$ \\
  $r_{11}\ =\ \dcall{m_7}{T_{\concept{x}}}{\module{Pickup}}{\X}{m_8}$ \\
  $r_{12}\ =\ \dcall{m_7}{\neg T_{\concept{x}}}{\module{Unstack}}{\X,\concept{B}}{m_8}$ \\
  $r_{13}\ =\ \dcall{m_8}{}{\module{Stack}}{\X,\Y}{m_8}$
\end{tabular}

\medskip
Rules $r_0$--$r_7$ are like the ones for the indexical policy $\pion^*$.
The external rules, however, are do-rules that apply ground actions to
remove blocks above \reg{0}, and to pick \X and put it on \Y.
The Boolean $T_{\concept{x}}$ is used to decide whether to use a \module{Pickup}
or \module{Unstack} action to grab \X. \hfill\eex

\medskip\noindent\textbf{Example 5.}
The module $\module{tower}(\role{O},\X)$ is aimed at the class \QTower of problems
where blocks are to be stacked into a \emph{single tower} resting on the table.
The goal of such a problem is described by a conjunction of atoms $\wedge_{i=1}^k on(x_{i},x_{i-1})$
and $ontable(x_0)$.
The module is the tuple \tup{\tup{\concept{O},\concept{X}},Z,M,\Reg,\Phi,m_0,R}
where \role{O} is a role argument whose denotation contains the pairs
$\{(x_i,x_{i-1}):i=1,...,k\}$, and \X is a concept argument that denotes the
lowest block in the target tower that is misplaced.\footnote{Block $x$ is \emph{well-placed}
  in state $s$ iff $on(x,y)$ holds in $s$ when $(x,y)$ is in \role{O}, and recursively %$x$ is on $y$ if the pair $(x,y)$ is in \role{O}, and recursively,
  $y$ is well-placed; it is \emph{misplaced} iff it is not well-placed. In the example,
  it is also assumed that the lowest block of the target tower must be placed on the table.
}
The other elements in the module are $Z{\,=\,}\emptyset$, $M{\,=\,}\{m_0,m_1,\ldots,m_3\}$,
$\Reg{\,=\,}\{\reg{0}\}$, and a set of features $\Phi=\{\concept{M},\concept{W}\}$  where
\concept{M} is the indexical concept that contains the block to be placed above the block
in \reg{0} according to \role{O}, if any, and, \concept{W} is the indexical concept that
contains the block directly below the block in \reg{0}, if any, also according to the target
tower \role{O}. The rules are:

\medskip
\begin{tabular}{@{}l}
  \textcolor{black}{\textit{\% Module {\normalfont$\module{tower}(\role{O},\X)$}}} \\%[2pt]
  $r_0\ =\ \nrule{m_0}{\GT{\concept{X}}}{\Load{\concept{X}}{\reg{0}},\UNK{\concept{M}},\UNK{\concept{W}}}{m_1}$ \\
  $r_1\ =\ \dcall{m_1}{\EQ{\concept{W}}}{\module{on-table}}{\reg{0}}{m_2}$ \\
  $r_2\ =\ \dcall{m_1}{\GT{\concept{W}}}{\module{on}}{\reg{0},\concept{W}}{m_2}$ \\
  $r_3\ =\ \dcall{m_2}{\GT{\concept{M}}}{\module{tower}}{\role{O},\concept{M}}{m_3}$
\end{tabular}

\medskip
The rule $r_0$ puts the lowest misplaced block in \reg{0}, $r_1$ calls the module
$\module{on-table}(\reg{0})$ to place \reg{0} on the table, $r_2$ calls the module
$\module{on}$ to place \reg{0} on \concept{W}, and $r_3$ \textbf{recursively calls}
the module with the block that is supposed to be on \reg{0}.
The module $\module{on-table}(\X)$, that puts the block in the singleton
\X on the table, is not spelled out.
\hfill\eex

\medskip\noindent\textbf{Example 6.}
The module $\module{blocks}(\role{O})$ solves arbitrary instances
of Blocksworld. It works by calling the module $\module{tower}(\role{O},\concept{X})$
with the parameter \X being the singleton that contains the \emph{lowest misplaced block}
in one of the current towers.
The module takes a single role argument \role{O} whose denotation encodes
the pairs $(x,y)$ corresponding to the goal atoms $on(x,y)$ as in Example~5.
The module is the tuple \tup{\tup{\concept{O}},Z,M,\Reg,\Phi,m_0,R} where
$Z = \emptyset$, $M=\{m_0,m_1\}$, $\Reg=\{r_0\}$, and $\Phi=\{\concept{L}\}$,
where \concept{L} contains the lowest misplaced blocks according to the
pairs in \role{O}.
The set $R$ contains only 2 rules:

\medskip
\begin{tabular}{@{}l}
  \textcolor{black}{\textit{\% Module {\normalfont$\module{blocks}(\role{O})$}}} \\
  $r_0\ =\ \nrule{m_0}{\GT{\concept{L}}}{\Load{\concept{L}}{\reg{0}}}{m_1}$ \\
  $r_1\ =\ \dcall{m_1}{}{\module{tower}}{\role{O},\reg{0}}{m_0}$
\end{tabular}

\medskip
The concept \concept{L} is a high-complexity concept, meaning that 
computing its denotation at the current state is non-trivial.
However, \concept{L} can be removed as follows.
First, a new marker \preg is put on a block $x$ such that the atoms
$on_G(x,y)$ and $on(x,y')$, with $y\neq y'$, hold at the current
state $s$; i.e., \preg is put on a misplaced block $x$.
Second, move the marker \preg to $y$ if $y$ is also misplaced,
and keep moving it down until the marked block is on a well-placed
block. At this moment, the marked block is a lowest misplaced block.
\hfill\eex

\subsection{Execution Model: \siwM}

The execution model for modules is captured by the \siwM algorithm
in Alg.~\ref{Xalg:siwM} which uses a stack and a caller/callee protocol,
as it is standard in programming languages.
It assumes  a collection $\{\module{mod}_0,\module{mod}_1,\ldots,\module{mod}_N\}$
of modules where the ``entry'' module $\module{mod}_0$ is assumed to take no arguments.
The execution  involves  solving classical planning subproblems, internal operations
on the registers, calls to other modules, and executions  of ground actions.
The modules do not share memory states nor registers, but they all act on the
(external) planning states $s$.

At each time point during execution, there is a single active module $\module{mod}_\ell$
that defines the current set of rules, and there is a current augmented state $(s,m,\v)$.
While no call or do rule is selected, \siwM behaves exactly as \siwRx.
However, if a call rule \call{m}{C}{\module{mod}_j}{x_1,x_2,\ldots,x_n}{m'} is chosen,
where $\module{mod}_j$ refers to \tup{\args,Z,M,\Reg,\Phi,m_0,R,}, the following
instructions are executed:
\begin{enumerate}[1.]\itemsep-1pt\topsep0pt
  \item Push context $(\ell,\v,m')$ where $\v$ is value for registers,
  \item Set value of arguments of $\module{mod}_j$ to those given by $x_i$,
  \item Set memory state to $m_0$ (the initial state of $\module{mod}_j$),
  \item Set the current set of rules to $R$,
  \item[5.] [Cont.\ execution of $\module{mod}_j$ until no rule is applicable], and
  \item[6.] Pop context $(\ell,\v,m')$, set value of registers to \v, memory state to $m'$,
    and rules $R$ to those in $\module{mod}_\ell$.
\end{enumerate}
If a do rule \call{m}{C}{\module{name}}{x_1,x_2,\ldots,x_n}{m'}
is chosen, on the other hand, then an applicable ground action $\module{name}(o_1,o_2,\ldots,o_n)$
at the current state $s$ with the object $o_i$ in $x_i$, $1\leq i\leq n$, is
applied and the memory state is set to $m'$. If no such ground action exists,
an error code is returned.
The implementation of the \siwM interpreter and all extended,
indexical policies and sketches, and modules that were discussed in the paper
are available online \cite{drexler-et-al-zenodo2024}.

%
% Dominik 2023-12-12: Anonymized for submission
% An implementation of the \siwRx algorithm on the Mimir planning system
% \cite{stahlberg-ecai2023}, together with the examples, is available
% \cite{drexler-et-al-zenodo2023c}.

\begin{algorithm}[t]
  \begin{algorithmic}[1]\small
    \algrenewcommand\algorithmicindent{0.75em}%
    \smallskip
    \State \textbf{Input:} Collection $\mathcal{M}=\{\module{mod}_0,\module{mod}_1,\ldots,\module{mod}_N\}$ of
      modules with entry module $\module{mod}_0$ % over the features in $\Phi$
    \State \textbf{Input:} Planning problem $P$ with initial state $s_0$ on which the features in $\Phi$ are well defined
    \State Initialize stack
    \State Let $\Reg^j$, $m^j_0$, and $R^j$ be the set of registers, initial memory, and rules of $\module{mod}_j$, $j=0,1,\ldots,N$
    \State $\ell\gets 0$, $R\gets R^\ell$, and $\bar s \gets (s_0,m^\ell_0,\v)$ for $\smash{\v\in\Obj(P)^{\Reg^\ell}}$
    \State\textbf{while} $s$ in $\bar s=(s,m,\v)$ is not a goal state of $P$ %\textbf{do}
    \State\quad\textbf{if} $m$ is internal memory %\textbf{then}
    \State\quad\quad Find rule $r=\xrule{(m,C)}{(E,m')}$ with $s,\v\vDash C$
    \State\quad\quad\textbf{if} $r$ is not found %\textbf{then}
    \State\quad\quad\quad\textbf{if} stack is empty, \Return FAILURE \hfill\textcolor{black}{\% Stalled}
    \State\quad\quad\quad Pop context $(j,\v',m')$ from stack
    \State\quad\quad\quad $\ell\gets j$, $R\gets R^\ell$, $m\gets m'$ and $\v \gets \v'$ %\hfill\textcolor{black}{\% back at $\module{mod}_j$}
    \State\quad\quad\textbf{else}
    \State\quad\quad\quad\textbf{if} $\Load{\C}{\preg}$ in $E$, $\v[\preg]\gets o$ for some $o\in\C(s,\v)$
    \State\quad\quad\quad $\bar s\gets(s,m',\v)$
    \State\quad\textbf{else} \hfill\textcolor{black}{\% $m$ is external memory}
    \State\quad\quad Find call/do rule $r{=}(m,C){\mapsto}(E,m')$ with $s,\v\vDash C$ %\xrule{(m,C)}{(E,m')}$ with $s,\v\vDash C$
    \State\quad\quad\textbf{if} $r=\call{m}{C}{\module{mod}_j}{x_1,x_2\ldots,x_n}{m'}$ %\textbf{then}
    \State\quad\quad\quad Push context $(\ell,\v,m')$ into stack
    \State\quad\quad\quad $R\gets R^j$, $m\gets m^j_0$ \hfill\textcolor{black}{\% Hand control to $\module{mod}_j$}
    \State\quad\quad\textbf{elsif} $r=\call{m}{C}{\module{name}}{x_1,x_2\ldots,x_n}{m'}$ %\textbf{then}
    \State\quad\quad\quad Find ground action $a=\module{name}(o_1,o_2,\ldots,o_n)$ appli- %cable
    \Statex\quad\quad\quad\quad cable at $s$ with $o_i\in x_i(s,\v)$, $i=1,2,\ldots,n$
    \State\quad\quad\quad\textbf{if} there is no such action, \Return FAILURE
    \State\quad\quad\quad $\bar s\gets(s',m',\v)$ where $(s,a,s')$ is transition in $P$
    \State\quad\quad\textbf{else} \hfill\textcolor{black}{\% No such rule is found}
    \State\quad\quad\quad Run IW search from $s$ to find goal state $s'$ of $P$, or
    \State\quad\quad\quad\quad state $s'$ such that $s'\prec_{r/\v} s$ for some (external) rule
    \State\quad\quad\quad\quad $r=\xrule{(m,C)}{(E,m')}$ in $R$
    \State\quad\quad\quad\textbf{if} no such state is found, \Return FAILURE
    \State\quad\quad\quad $\bar s \gets (s',m',\v)$
    \State\Return path from $s_0$ to the goal state $s$
  \end{algorithmic}
  \caption{\siwM uses set of modules $M$ (extended sketches) for solving
    a problem $P$ via possibly nested calls, execution of ground actions,
    and IW searches.
  }
  \label{Xalg:siwM}
\end{algorithm}

\section{Discussion}
\label{sect:conclusions}
%\vskip -.5em

A basic, concrete question about policy reuse in a planning setting is: can a policy for putting one block
on top of another be used for building any  given tower of blocks, and eventually any block configuration?
The question is relevant because it tells us that policies for building given towers and block configurations
do not have to be learned from scratch, but that they can be learned bottom up, from simpler to  complex,
one after the other \cite{dreamer}. The subtlety is that  in order for complex policies to use simpler policies,
the former must pass the right  parameters to the latter depending on the context defined by the top goal and the
current state. In this work, we have developed a language for representing such  policies and sketches, and this form of hierarchical composition.

We have also addressed another source of complexity when learning general policies and sketches:
the complexity of the features involved. We have shown that indexical features whose values depend on
the content of  registers that can be dynamically loaded with objects, can be used to drastically reduce
the complexity of the features needed, in line with the intuition of the so-called deictic representations,
where a constant number of ``visual marks'' are used to mark objects so that they can be easily referred to
\cite{chapman:penguins,ballard-et-al-bbs1996,finney-et-al-arxiv2013}. In our setting, an object can be regarded
as marked with ``mark'' \preg when the object is loaded into register \preg.

The use of registers and concept features allows for general policies that map states into
ground actions too, even if the set of ground actions change from instance to instance. This is different
than general policies defined as filters on state transitions \cite{bonet-geffner-ijcai2018,frances-et-al-aaai2021},
which require a model for determining the possible transitions from a state.
Policies that map states into actions are more conventional and can be applied model-free.

The achieved expressivity is the result of three extensions in the  language of policies and sketches:
internal memory states,  like in finite state controllers for sequencing behaviors, indexical concepts
and features, whose denotation depends on the value of registers that can be updated,
and modules that wrap up policies and sketches and allow them to call each other
by passing parameters as a function of the state, the registers, and the goals (represented in the state).

The language of extended policies and sketches adds an interface
for calling policies and sketches from other policies and sketches, even recursively,
as illustrated in the examples.
The resulting language has elements in common with
programming languages and planning programs \cite{javi1,javi2,javi3},
but there are key differences too.
In particular, the use of a rich feature language does not limit policies and sketches
to deal  with classes of problems with a fixed goal and the
policy and sketch modules do not have to represent full procedures or policies;
they can also represent sketches where ``holes'' are filled in with polynomial
IW searches  when the sketches have bounded width.

Provided with this richer language for policies and sketches,
the next step is learning them from  small problem instances,
adapting the  methods developed for learning policies and  sketches \cite{frances-et-al-aaai2021,drexler-et-al-icaps2022}.
The new, extended language also opens the possibility of 
learning  hierarchical policies bottom-up by generating and reusing policies,
instead of the more common approach of learning them top-down \cite{drexler-et-al-kr2023}.
% , and to study the theoreticalproperties of the modular framework. % like acyclicity and termination.
Finally, the study of theoretical properties of the modular framework, like acyclicity
and termination, is also important and left for future work.

\section*{Acknowledgments}

The research of H. Geffner  has been supported by the Alexander von Humboldt Foundation with funds from
the Federal Ministry for Education and Research. It has also  received funding from
the European Research Council (ERC), Grant agreement No. No 885107, and Project TAILOR,
Grant agreement No. 952215,  under EU Horizon 2020 research and innovation programme,
the Excellence Strategy of the Federal Government and the NRW L\"{a}nder, Germany, and 
the Knut and Alice Wallenberg (KAW) Foundation under the WASP program.

%\input{chapters/acknowledgements.tex}

%\small
\bibliography{control,extra,bib/abbrv-short,bib/literatur,bib/crossref-short}
%{\bibliography{control,extra,bib/abbrv-short,bib/literatur}}%,bib/crossref-short}}
%\bibliographystyle{plainnat}

\end{document}

\maketitle

\section{Appendix}
\appendix

\begin{boxed-example}[General policy for Towers of Hanoi with 3 pegs]
  \label{ex:hanoi}
  \citet{hanoi:aaai2023} describe  a simple policy for the class \QHanoi of Towers-of-Hanoi problems
  with 3 pegs.  The policy is expressed by referring to the relative size of the disk
  being moved, among the top disks at each peg, either a movement of the
  smallest disk, or a movement of the other disk:
  %smallest disk (2 such possible movements), or a movement of the other disk (one such possible movement):
  \begin{quote}
    \it
    Alternate actions between the smallest disk and a non-smallest disk.
    When moving the smallest disk, always move it to the left.
    (If it is in the first peg, move it to the third peg.)
    If the smallest disk is on the first pillar, move it to the third one.
    When moving the non-smallest disk, take the only valid action.
  \end{quote}
  The movements are expressed in the language of sketches with
  three Boolean features $p_{i,j}$, $1 \leq i < j \leq 3$, that are true
  if the top disk at peg $i$ is smaller than the top disk at peg $j$.
  For example, the smallest disk is at peg 1 (resp.\ peg 3) iff
  $p_{1,2}\land p_{1,3}$ (resp.\ $\neg p_{1,3} \land \neg p_{2,3}$) holds.
  The alternation of movements is obtained by using two memory states $m_0$ and
  $m_1$ of which $m_0$ is the initial memory.
  The rules are written slightly different for clarity where a rule \xrule{(m,C)}{(E,m')}
  is expressed as \nrule{m}{C}{E}{m'}:

  \begin{xltabular}{\linewidth}{@{\ \ \ }lcX}
    %\textcolor{NavyBlue}{\it\% Movements of the smallest disk} \\[2pt]
    \nrule{m_0}{p_{1,2},p_{1,3}}{\UNK{p_{1,2}},\neg p_{1,3}, \neg p_{2,3}}{m_1}      &\qquad\quad & Move smallest disk from peg 1 to peg 3 \\[2pt]
    \nrule{m_0}{\neg p_{1,2},p_{2,3}}{p_{1,2},p_{1,3},\UNK{p_{2,3}}}{m_1}            &\qquad\quad & Move smallest disk from peg 2 to peg 1 \\[2pt]
    \nrule{m_0}{\neg p_{1,3}, \neg p_{2,3}}{\neg p_{1,2},\UNK{p_{1,3}},p_{2,3}}{m_1} &\qquad\quad & Move smallest disk from peg 3 to peg 2 \\[1em]
    %
    %\textcolor{NavyBlue}{\it\% Movements of the other disk} \\[2pt]
    \nrule{m_1}{p_{1,2},p_{1,3},p_{2,3}}{\neg p_{2,3}}{m_0}                          &\qquad\quad & Move other disk from peg 2 to peg 3 \\[2pt]
    \nrule{m_1}{p_{1,2},p_{1,3},\neg p_{2,3}}{p_{2,3}}{m_0}                          &\qquad\quad & Move other disk from peg 3 to peg 2 \\[2pt]
    \nrule{m_1}{\neg p_{1,2},p_{1,3},p_{2,3}}{\neg p_{1,3}}{m_0}                     &\qquad\quad & Move other disk from peg 1 to peg 3 \\[2pt]
    \nrule{m_1}{\neg p_{1,2},\neg p_{1,3},p_{2,3}}{p_{1,3}}{m_0}                     &\qquad\quad & Move other disk from peg 3 to peg 1 \\[2pt]
    \nrule{m_1}{p_{1,2},\neg p_{1,3},\neg p_{2,3}}{\neg p_{1,2}}{m_0}                &\qquad\quad & Move other disk from peg 1 to peg 2 \\[2pt]
    \nrule{m_1}{\neg p_{1,2},\neg p_{1,3},\neg p_{2,3}}{p_{1,2}}{m_0}                &\qquad\quad & Move other disk from peg 2 to peg 1
  \end{xltabular}
\end{boxed-example}

%\small
{\bibliography{control,extra,bib/abbrv-short,bib/literatur,bib/crossref-short}}
%{\bibliography{control,extra,bib/abbrv-short,bib/literatur}}%,bib/crossref-short}}
\bibliographystyle{plainnat}